\documentclass[10pt,twocolumn,letterpaper]{article}

\usepackage{cvpr}
\usepackage{times}
\usepackage{epsfig}
\usepackage{graphicx}
\usepackage{amsmath}
\usepackage{amssymb}
\usepackage{booktabs}
\usepackage{subfigure}
\newcommand{\tabincell}[2]{\begin{tabular}{@{}#1@{}}#2\end{tabular}}

\newcommand*{\affaddr}[1]{#1} 
\newcommand*{\affmark}[1][*]{\textsuperscript{#1}}
\newcommand*{\email}[1]{\texttt{#1}}

\usepackage[pagebackref=true,breaklinks=true,letterpaper=true,colorlinks,bookmarks=false]{hyperref}

\cvprfinalcopy 

\def\cvprPaperID{2452} 

\ifcvprfinal\fi
\begin{document}

\title{Siamese Box Adaptive Network for Visual Tracking}

\author{
Zedu Chen\affmark[1], Bineng Zhong\affmark[1, 6]\thanks{Corresponding author.}, Guorong Li\affmark[2], Shengping Zhang\affmark[3,4], Rongrong Ji\affmark[5,4]\\
\affaddr{\affmark[1]Department of Computer Science and Technology, Huaqiao University}\\
\affaddr{\affmark[2]School of Computer Science and Technology, University of Chinese Academy of Sciences}\\
\affaddr{\affmark[3]Harbin Institute of Technology},
\affaddr{\affmark[4]Peng Cheng Laboratory}\\
\affaddr{\affmark[5]Department of Artificial Intelligence, School of Informatics, Xiamen University}\\
\affaddr{\affmark[6]Key Laboratory of Intelligent Perception and Systems for High-Dimensional Information of\\ Ministry of Education, Nanjing University of Science and  Technology}\\
\email{zeduchen@stu.hqu.edu.cn},
\email{bnzhong@hqu.edu.cn},
\email{liguorong@ucas.ac.cn}\\
\email{s.zhang@hit.edu.cn},
\email{rrji@xmu.edu.cn}
}


\maketitle
\thispagestyle{empty}

\begin{abstract}
   Most of the existing trackers usually rely on either a multi-scale searching scheme or pre-defined anchor boxes to accurately estimate the scale and aspect ratio of a target. Unfortunately, they typically call for tedious and heuristic configurations. To address this issue, we propose a simple yet effective visual tracking framework (named Siamese Box Adaptive Network, SiamBAN) by exploiting the expressive power of the fully convolutional network (FCN). SiamBAN views the visual tracking problem as a parallel classification and regression problem, and thus directly classifies objects and regresses their bounding boxes in a unified FCN. The no-prior box design avoids hyper-parameters associated with the candidate boxes, making SiamBAN more flexible and general. Extensive experiments on visual tracking benchmarks including VOT2018, VOT2019, OTB100, NFS, UAV123, and LaSOT demonstrate that SiamBAN achieves state-of-the-art performance and runs at 40 FPS, confirming its effectiveness and efficiency. The code will be available at \url{https://github.com/hqucv/siamban}.
\end{abstract}

\section{Introduction}


\begin{figure}[t]
\begin{center}
\subfigure[Different scale or aspect ratio handing methods]{
\includegraphics[width=1\linewidth]{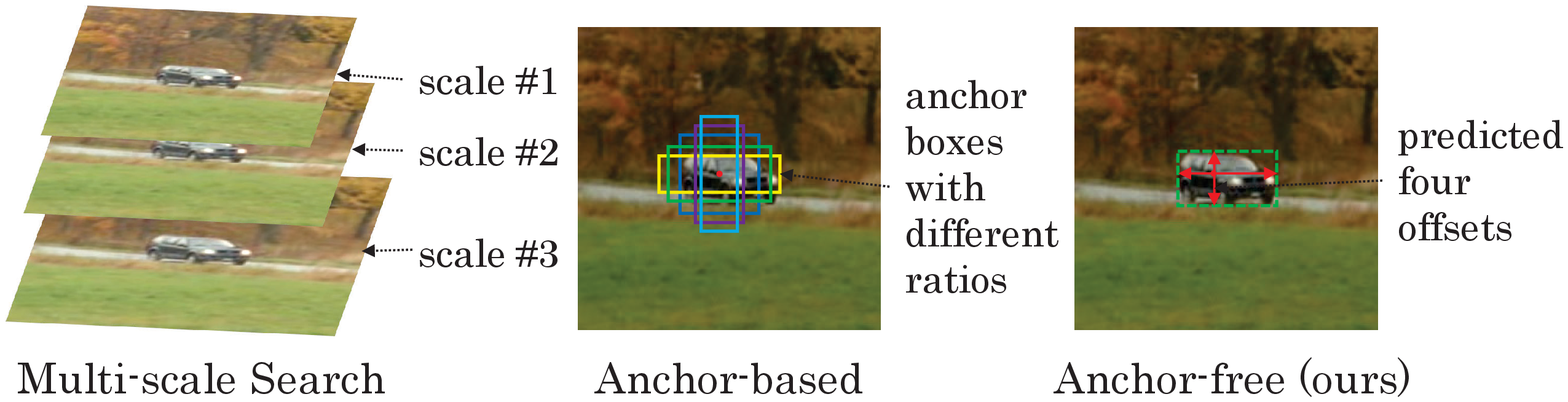}}
\subfigure[Qualitative comparison results]{
\includegraphics[width=1\linewidth]{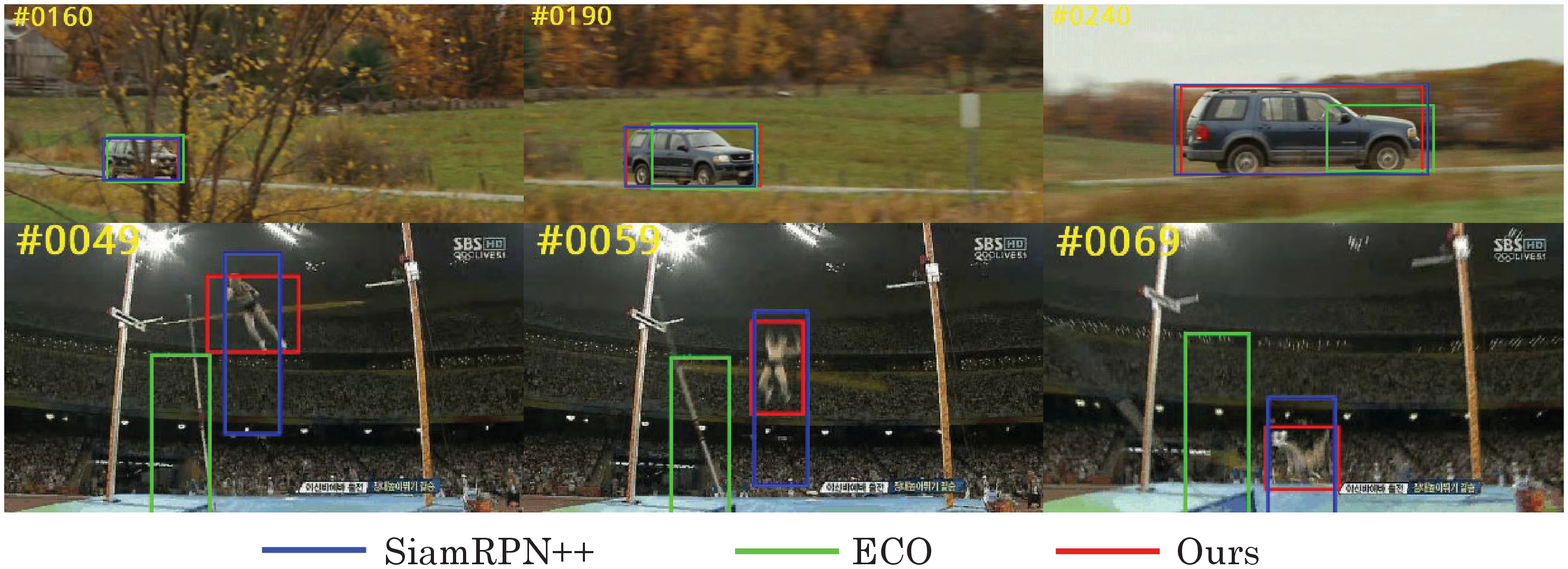}}
\end{center}
\caption{(a) Methods used to estimate the target scale or aspect ratio: multi-scale search (such as SiamFC, ECO), anchor-based (such as SiamRPN, SiamRPN++), and anchor-free (such as ours).(b) Some representative experiment results from our SiamBAN tracker and two state-of-the-art trackers. Observed from the visualization results, our tracker is better than the other two trackers in terms of scale and aspect ratio.}
\label{fig:diea}
\end{figure}

Visual tracking is a fundamental but challenging task in computer vision. Given the target state in the initial frame of a sequence, the tracker needs to predict the target state in each subsequent frame. Despite great progress in recent years, visual tracking still faces challenges due to occlusion, scale variation, background clutters, fast motion, illumination variation, and appearance variations.  


In a real-world video, the target scale and aspect ratio are also changing due to target or camera movement and target appearance changes. Accurately estimating the scale and aspect ratio of the target becomes a challenge in the field of visual tracking. However, many existing trackers ignore this problem and rely on a multi-scale search to estimate the target size. For example, the current state-of-the-art correlation filter based trackers \cite{ECO, UPDT} rely on their classification components, and the target scale is simply estimated by multi-scale search. Recently, Siamese network based visual trackers \cite{SiamRPN, DaSiamRPN, SiamRPN++} introduce a region proposal network (RPN) to obtain accurate target bounding boxes. However, in order to handle different scales and aspect ratios, they need to carefully design anchor boxes based on heuristic knowledge, which introduces many hyper-parameters and computational complexity.

In contrast, neuroscientists have shown that the bio-visual primary visual cortex can quickly and effectively extract the contours or boundaries of the observed objects from complex environments \cite{Neuroscience}. That is to say, humans can identify the object position and boundary without candidate boxes. So can we design an accurate and robust visual tracking framework without relying on candidate boxes? Inspired by the anchor-free detectors \cite{DenseBox, UnitBox, YOLOv1, FSAF, FCOS}, the answer is yes. By exploiting the expressive power of the fully convolutional network (FCN), we propose a simple yet effective visual tracking framework named Siamese box adaptive network (SiamBAN) to address the challenge of accurately estimating the scale and aspect ratio of the target. The framework consists of a Siamese network backbone and multiple box adaptive heads, which does not require pre-defined candidate boxes and can be optimized end-to-end during training. SiamBAN classifies the target and regresses bounding boxes directly in a unified FCN, transforming the tracking problem into a classification-regression problem. Specifically, it directly predicts the foreground-background category score and a 4D vector of each spatial position on the correlation feature maps. The 4D vector depicts the relative offset from the four sides of the bounding box to the center point of the feature location corresponding to the search region. During inference, we use a search image centered on the previous position of the target. Through the bounding box corresponding to the position of the best score, we can get the displacement and size change of the target between frames. 

The main contributions of this work are threefold. 
\begin{itemize}
\item We design a Siamese box adaptive network, which can perform end-to-end offline training with deep convolutional neural networks \cite{ResNet} on well-annotated datasets \cite{ImageNet2015, YoutubeBB, COCO, GOT10k, LaSOT}. 
\item The no-prior box design in SiamBAN avoids hyper-parameters associated with the candidate boxes, making our tracker more flexible and general.
\item SiamBAN not only achieves state-of-the-art results, but also runs at 40 FPS on tracking benchmarks including VOT2018 \cite{VOT2018}, VOT2019 \cite{VOT2019}, OTB100 \cite{OTB100}, NFS \cite{NFS}, UAV123 \cite{UAV123}, and LaSOT \cite{LaSOT}.
\end{itemize}

\section{Related Works}

Visual tracking is one of the most active research topics in computer vision in recent decades. A comprehensive survey of the related trackers is beyond the scope of this paper, so we only briefly review two aspects that are most relevant to our work: Siamese network based visual trackers and anchor-free object detectors.

\subsection{Siamese Network Based Visual Trackers}

Recently, Siamese network based trackers have attracted great attention from the visual tracking community due to their end-to-end training capabilities and high efficiency \cite{SiamFC, DSiam, RASNet, SiamRPN, SiamRPN++, SiamDW}. SiamFC \cite{SiamFC} adopts the Siamese network as a feature extractor and first introduces the correlation layer to combine feature maps. Owing to its light structure and no need to model update, SiamFC runs efficiently at 86 FPS. DSiam \cite{DSiam} learns a feature transformation to handle the target appearance variation and to suppress background. RASNet \cite{RASNet} embeds diverse attention mechanisms in the Siamese network to adapt the tracking model to the current target. However, these methods need a multi-scale test to cope with scale variation and cannot handle aspect ratio changes due to target appearance variations. In order to get a more accurate target bounding box, SiamRPN \cite{SiamRPN} introduces the RPN \cite{Faster_r-cnn} into the SiamFC. SPM-Tracker \cite{SPM-Tracker} proposes a series-parallel matching framework to enhance the robustness and discrimination power of SiamRPN. SiamRPN++ \cite{SiamRPN++}, SiamMask \cite{SiamMask} and SiamDW \cite{SiamDW} remove the influence factors such as padding in different ways, and introduce modern deep neural networks such as ResNet \cite{ResNet}, ResNeXt \cite{ResNeXt} and MobileNet \cite{MobileNet} into the Siamese network based visual trackers. Although anchor-based trackers \cite{SiamRPN, SPM-Tracker, SiamRPN++} can handle changes in scale and aspect ratio, it is necessary to carefully design and fix the parameters of the anchor boxes. Design parameters often requires heuristic adjustments and involves many tricks to achieve good performance. In contrast to anchor-based trackers, our tracker avoids hyper-parameters associated with the anchor boxes and is more flexible and general.

\subsection{Anchor-free Object Detectors}

\begin{figure*}
\begin{center}
   \includegraphics[width=1\linewidth]{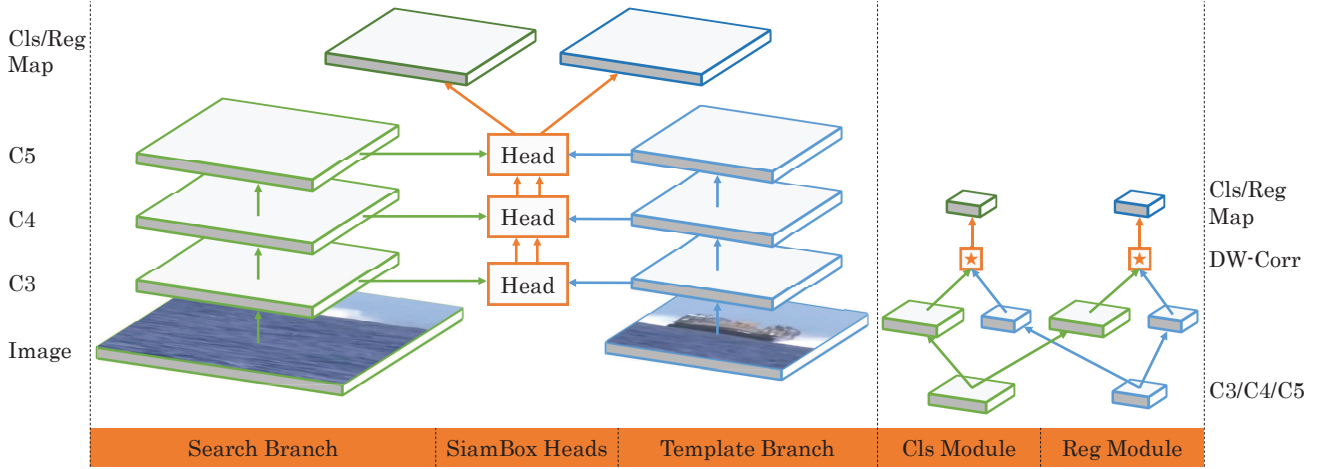}
\end{center}
   \caption{The framework of the proposed Siamese box adaptive network. The left sub-figure shows its main structure, where C3, C4, and C5 denote the feature maps of the backbone network, Cls Map and Reg Map denote the feature maps of the SiamBAN heads output. The right sub-figure shows each SiamBAN head, where DW-Corr means depth-wise cross-correlation operation.}
\label{fig:siamban}
\end{figure*}

Recently, anchor-free object detection has attracted the attention of the object detection community. However, anchor-free detection is not a new concept. DenseBox \cite{DenseBox} first introduced an FCN framework to jointly perform face detection and landmark localization. UnitBox \cite{UnitBox} offered another option for performance improvement by carefully designing optimization losses. YOLOv1 \cite{YOLOv1} proposed to divide the input image into a grid and then predicted bounding boxes and class probabilities on each grid cell. 

Recently, many new anchor-free detectors have emerged. These detection methods can be roughly classified into keypoint based object detection \cite{CornerNet, ExtremeNet, RepPoints} and dense detection \cite{FSAF, FCOS}. Specifically, CornerNet \cite{CornerNet} proposed to detect an object bounding box as a pair of keypoints. ExtremeNet \cite{ExtremeNet} presented to detect four extreme points and one center point of objects using a standard keypoint estimation network. RepPoints \cite{RepPoints} introduced the representative points, a new representation of objects to model fine-grained localization information and identify local areas significant for object classification. FSAF \cite{FSAF} proposed feature selective anchor-free module to address the limitations imposed by heuristic feature selection for anchor-based single-shot detectors with feature pyramids. FCOS \cite{FCOS} proposed to directly predict the possibility of object existence and the bounding box coordinates without anchor reference.

Compared to object detection, there are two key challenges in the visual tracking task, i.e. unknown categories and discrimination between different objects. The anchor-free detectors usually assume the categories of the objects to be detected are pre-defined. However, the categories of the targets are unknown before tracking. Meanwhile, anchor-free detectors typically focus on detecting the objects from different categories, while in tracking, it is necessary to determine whether the two objects are the same one. Therefore, a template branch that can encode the appearance information is need in our framework to identify the target and background.


\section{SiamBAN Framework}

In this section, we describe the proposed SiamBAN framework. As shown in Figure \ref{fig:siamban}, SiamBAN consists of a Siamese network backbone and multiple box adaptive heads. The Siamese network backbone is responsible for computing the convolutional feature maps of the template patch and the search patch, which uses an off-the-shelf convolutional network. The box adaptive head includes a classification module and a regression module. Specifically, the classification module performs foreground-background classification on each point of the correlation layer, and the regression module performs bounding box prediction on the corresponding position.

\subsection{Siamese Network Backbone}


Modern deep neural networks \cite{ResNet, ResNeXt, MobileNet} have proven to be effective in Siamese network based trackers \cite{SiamRPN++, SiamMask, SiamDW}, and now we can use them such as ResNet, ResNeXt, and MobileNet in Siamese network based trackers. In our tracker, we adopt ResNet-50 \cite{ResNet} as the backbone network. Although ResNet-50 with continuous convolution striding can learn more and more abstract feature representations, it reduces feature resolution. However, Siamese network based trackers need detailed spatial information to perform dense predictions. To deal with this problem, we remove the downsampling operations from the last two convolution blocks. At the same time, in order to improve the receptive field, we use atrous convolution \cite{DeepLabv2}, which is proven to be effective for visual tracking \cite{SiamRPN, SiamMask}. In addition, inspired by multi-grid methods \cite{DUC}, we adopt different atrous rates in our model. Specifically, we set the stride to 1 in the $conv4$ and $conv5$ blocks, the atrous rate to 2 in the $conv4$ block, and the atrous rate to 4 in the $conv5$ block.


The Siamese network backbone consists of two identical branches. One is called the template branch, which receives the template patch as input (denoted as $z$). The other is called the search branch, which receives the search patch as input (denoted as $x$). The two branches share parameters in a convolutional neural network to ensure that the same transformation is applied to both patches. In order to reduce the computational burden, we add a $1 \times 1$ convolution to reduce the output features channel to 256, and use only the features of the template branch center $7 \times 7$ regions \cite{CFNet, SiamRPN++}, which can still capture the entire target region. For convenience, the output features of the Siamese network are represented as $\varphi(z)$ and $\varphi(x)$.

\subsection{Box Adaptive Head}

As shown in Figure \ref{fig:siamban} (right), box adaptive head consists of a classification module and a regression module. Both modules receive features from the template branch and the search branch. So we adjust and copy $\varphi(z)$ and $\varphi(x)$ to $[\varphi(z)]_{cls}$, $[\varphi(z)]_{reg}$ and $[\varphi(x)]_{cls}$, $[\varphi(x)]_{reg}$ to the corresponding module. According to our design, each point of the correlation layer of the classification module needs to output two channels for foreground-background classification, and each point of the correlation layer of the regression module needs to output four channels for prediction of the bounding box. Each module combines the feature maps using a depth-wise cross-correlation layer \cite{SiamRPN++}:
\begin{equation}
\begin{split}
    &P_{w \times h \times 2}^{cls} = [\varphi(x)]_{cls} \star [\varphi(z)]_{cls}, \\
    &P_{w \times h \times 4}^{reg} = [\varphi(x)]_{reg} \star [\varphi(z)]_{reg},
\label{eq:cross-correlation}
\end{split}
\end{equation}
where $\star$ denotes the convolution operation with $[\varphi(z)]_{cls}$ or $[\varphi(z)]_{reg}$ as the convolution kernel, $P_{w \times h \times 2}^{cls}$ denotes classification map, $P_{w \times h \times 4}^{reg}$ indicates regression map. It is worth noting that our tracker outputs 5 times fewer variables than anchor-based trackers \cite{SiamRPN, SiamRPN++} with 5 anchor boxes. 

For each location on the classification map $P_{w \times h \times 2}^{cls}$ or the regression map $P_{w \times h \times 4}^{reg}$, we can map it to the input search patch. For example, the location $(i,\ j)$ corresponding to the location on the search patch is $[\lfloor \frac{w_{im}}{2} \rfloor + (i - \lfloor \frac{w}{2} \rfloor) \times s,\ \lfloor \frac{h_{im}}{2} \rfloor + (j - \lfloor \frac{h}{2} \rfloor) \times s]$ (denoted as $(p_i,\ p_j)$. $w_{im}$ and $h_{im}$ represent the width and height of the input search patch and $s$ represents the total stride of the network), which is the center of the receptive field of the position $(i,\ j)$. For the regression, the anchor-based trackers \cite{SiamRPN, DaSiamRPN, SiamRPN++} treat the location $(p_i,\ p_j)$ as the center of the anchor box and regress the location $(p_i,\ p_j)$, width $a_w$ and height $a_h$. That is, for the position $(i,\ j)$, the regression can adjust all of its offset values, but the classification is still performed in the original position, which may result in inconsistencies in classification and regression. So we do not adjust the location $(p_i,\ p_j)$ and only calculate its offset value to the bounding box. In addition, since our regression targets are positive real numbers,  we apply $exp(x)$ at the last level of the regression module to map any real number to $(0,\ +\infty)$. 


\subsection{Multi-level Prediction}

After utilizing ResNet-50 with atrous convolution, we can use multi-level features for prediction. Although the spatial resolutions of the $conv3$, $conv4$ and $conv5$ blocks of our backbone network are the same, they have atrous convolutions with different expansion rates, so the difference between their receptive fields is large, and the captured information is naturally different. As pointed by CF2 \cite{CF2}, features from earlier layers can capture fine-grained information, which is useful for precise localization; while features from latter layers can encode abstract semantic information, which is robust to target appearance changes. In order to take full advantage of different characteristics of multi-level features, we use multiple box adaptive heads for prediction. The classification maps and the regression maps obtained by each detection head are adaptively fused:
\begin{equation}
\begin{split}
    &P_{w \times h \times 2}^{cls-all} = \sum_{l=3}^{5} \alpha_l P_{l}^{cls}, \\
    &P_{w \times h \times 4}^{reg-all} = \sum_{l=3}^{5} \beta_l P_{l}^{reg}, \label{multi-level}
\end{split}
\end{equation}
where $\alpha_l$ and $\beta_l$ are the weights corresponding to each map and are optimized together with the network. By combining the classification map and the regression map independently, the classification module and the regression module can focus on the domains they need.

\subsection{Ground-truth and Loss}\label{sec:ground-truth and loss}

\begin{figure}[t]
\begin{center}
   \includegraphics[width=1\linewidth]{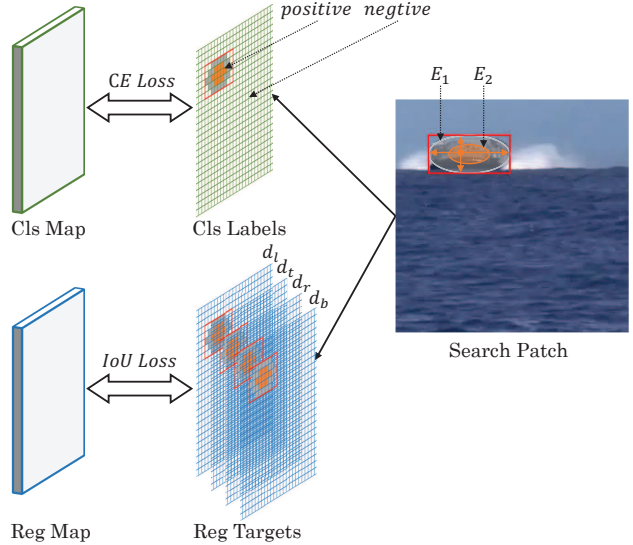}
\end{center}
   \caption{Illustrations of classification labels and regression targets. Prediction values and supervision signals are as shown, where $E_1$ represents ellipse $E_1$ and $E_2$ represents ellipse $E_2$. We use a cross entropy and an IoU loss for classification and box regression, respectively.}
\label{fig:groundtruth}
\end{figure}

\noindent
\textbf{Classification Labels and Regression Targets.} \ As shown in Figure \ref{fig:groundtruth}, the target on each search patch is marked with a ground-truth bounding box. The width, height, top-left corner, center point and bottom-right corner of the ground-truth bounding box are represented by $g_w$, $g_h$, $(g_{x_1},\ g_{y_1})$, $(g_{x_c},\ g_{y_c})$ and $(g_{x_2},\ g_{y_2})$, respectively. With $(g_{x_c},\ g_{y_c})$ as the center and $\frac{g_w}{2}$, $\frac{g_h}{2}$ as the axes length, we can get the ellipse $E_1$:
\begin{equation}
    \frac{(p_i - g_{x_c})^2}{(\frac{g_w}{2})^2} + \frac{(p_j - g_{y_c})^2}{(\frac{g_h}{2})^2} = 1.
\label{eq:ellipse1}
\end{equation}
With $(g_{x_c},\ g_{y_c})$ as the center and $\frac{g_w}{4}$, $\frac{g_h}{4}$ as the axes length, we can get the ellipse $E_2$:
\begin{equation}
    \frac{(p_i - g_{x_c})^2}{(\frac{g_w}{4})^2} + \frac{(p_j - g_{y_c})^2}{(\frac{g_h}{4})^2} = 1.
\label{eq:ellipse2}
\end{equation}
If the location $(p_i,\ p_j)$ falls within the ellipse $E_2$, it is assigned with a positive label, and if it falls outside the ellipse $E_1$, it is assigned with a negative label, and if it falls between the ellipses $E_2$ and $E_1$, ignore it. The location $(p_i,\ p_j)$ with a positive label is used to regress the bounding box, and the regression targets can be formulated as:
\begin{equation}
\begin{split}
    d_l = p_i - g_{x_1}, \\
    d_t = p_j - g_{y_1}, \\
    d_r = g_{x_2} - p_i, \\
    d_b = g_{y_2} - p_j,
\label{eq:regression}
\end{split}
\end{equation}
where $d_l$, $d_t$, $d_r$, $d_b$ are the distances from the location to the four sides of the bounding box, as shown in Figure \ref{fig:groundtruth}.

\noindent
\textbf{Classification Loss and Regression Loss.} \ We define our multi-task loss function as follows:
\begin{equation}
    L = \lambda_1 L_{cls} + \lambda_2 L_{reg},
\label{eq:loss}
\end{equation}
where $L_{cls}$ is the cross entropy loss, $L_{reg}$ is the IoU (Intersection over Union) Loss. We do not search for the hyper-parameters of Eq.\ref{eq:loss}, and simply set $\lambda_1 = \lambda_2 = 1$. Similar to GIoU \cite{GIoU}, we define IoU loss as:
\begin{equation}
    L_{IoU} = 1 - IoU,
\end{equation}
where $IoU$ represents the area ratio of intersection to union of the predicted bounding box and the ground-truth bounding box. The location $(p_i,\ p_j)$ with a positive label is within the ellipse $E_2$ and the regression value is greater than 0, so $0 < IoU \leq 1$, then $0 \leq L_{IoU} < 1$. The IoU loss can make $d_l$, $d_t$, $d_r$, $d_b$ jointly be regressed.

\subsection{Training and Inference}

\noindent
\textbf{Training.} \ Our entire network can be trained end-to-end on large-scale datasets. We train SiamBAN with image pairs sampled on videos or still images. The training sets include ImageNet VID \cite{ImageNet2015}, YouTube-BoundingBoxes \cite{YoutubeBB}, COCO \cite{COCO}, ImageNet DET \cite{ImageNet2015}, GOT10k \cite{GOT10k} and LaSOT \cite{LaSOT}. The size of a template patch is $127 \times 127$ pixels, while the size of a search patch is $255 \times 255$ pixels. Also, although our negative samples are much less than anchor-based trackers \cite{SiamRPN, SiamRPN++}, negative samples are still much more than positive samples. Therefore we collect at most 16 positive samples and 48 negative samples from one image pair.

\noindent
\textbf{Inference.} \ During inference, we crop the template patch from the first frame and feed it to the feature extraction network. The extracted template features are cached, so we do not have to calculate them in subsequent tracking. For subsequent frames, we crop the search patch and extract feature based on the target position of the previous frame, and then perform prediction in the search region to get the total classification map $P_{w \times h \times 2}^{cls-all}$ and regression map $P_{w \times h \times 2}^{reg-all}$. Afterward, we can get prediction boxes by the following equation:
\begin{equation}
\begin{split}
    p_{x_1} = p_i - d_l^{reg}, \\
    p_{y_1} = p_j - d_t^{reg}, \\
    p_{x_2} = p_i + d_r^{reg}, \\
    p_{y_2} = p_j + d_b^{reg},
\label{eq:prediction}
\end{split}
\end{equation}
where $d_l^{reg}$, $d_t^{reg}$, $d_r^{reg}$ and $d_b^{reg}$ denote the prediction values of the regression map, $(p_{x_1},\ p_{y_1})$ and $(p_{x_2},\ p_{y_2})$ are the top-left corner and bottom-right corner of the prediction box.

After prediction boxes are generated, we use the cosine window and scale change penalty to smooth target movements and changes \cite{SiamRPN}, then the prediction box with the best score is selected and its size is updated by linear interpolation with the state in the previous frame.

\section{Experiments}

\subsection{Implementation Details}
We initialize our backbone networks with the weights pre-trained on ImageNet \cite{ImageNet2015} and the parameters of the first two layers are frozen. Our network is trained with stochastic gradient descent (SGD) with a minibatch of 28 pairs. We train a total of 20 epochs, using a warmup learning rate of 0.001 to 0.005 in the first 5 epochs and a learning rate exponentially decayed from 0.005 to 0.00005 in the last 15 epochs. In the first 10 epochs, we only train the box adaptive heads, and in the last 10 epochs fine-tuned the backbone network with one-tenth of the current learning rate. Weight decay and momentum are set as 0.0001 and 0.9. Our approach is implemented in Python using PyTorch on a PC with Intel Xeon(R) 4108 1.8GHz CPU, 64G RAM, Nvidia GTX 1080Ti. 

\subsection{Comparison with State-of-the-art Trackers}

We compare our \textbf{SiamBAN} tracker with the state-of-the-art trackers on six tracking benchmarks. Our tracker achieves state-of-the-art results and run at 40 FPS.

\begin{table*}[ht]
\begin{center}
\setlength{\tabcolsep}{2.4mm}{
\begin{tabular}{lcccccccccc}
\toprule
& \tabincell{c}{DRT \\ \cite{DRT}}  & \tabincell{c}{RCO \\ \cite{VOT2018}}  & \tabincell{c}{UPDT \\ \cite{UPDT}}  & \tabincell{c}{SiamRPN \\ \cite{SiamRPN}}  & \tabincell{c}{MFT \\ \cite{VOT2018}}  & \tabincell{c}{LADCF \\ \cite{LADCF}}  & \tabincell{c}{ATOM \\ \cite{ATOM}}  & \tabincell{c}{SiamRPN++ \\ \cite{SiamRPN++}}  & \tabincell{c}{DiMP \\ \cite{DiMP}}  & \textbf{Ours}  \\
\midrule
EAO$\uparrow$          & 0.355 & 0.376 & 0.379 & 0.384   & 0.386 & 0.389 & 0.401 & 0.417     & {\color{blue} 0.441} & {\color{red} 0.452} \\
Accuracy$\uparrow$     & 0.518 & 0.507 & 0.536 & 0.588   & 0.505 & 0.503 & 0.590 & {\color{red} 0.604}     & {\color{blue} 0.597} & {\color{blue} 0.597} \\
Robustness$\downarrow$ & 0.201 & 0.155 & 0.184 & 0.276   & {\color{red} 0.140} & 0.159 & 0.203 & 0.234     & {\color{blue} 0.152} & 0.178 \\
\bottomrule
\end{tabular}}
\end{center}
\caption{Detailed comparisons on VOT2018. The best two results are highlighted in {\color{red} red} and {\color{blue} blue} fonts. DiMP is the ResNet-50 version (DiMP-50), the same below.}
\label{tab:eao_rank_vot2018}
\end{table*}

\begin{table*}[ht]
\begin{center}
\setlength{\tabcolsep}{0.95mm}{
\begin{tabular}{lcccccccccc}
\toprule
& \tabincell{c}{SA\_SIAM\_R \\ \cite{VOT2019}} & \tabincell{c}{SiamCRF\_RT \\ \cite{VOT2019}} & \tabincell{c}{SPM \\ \cite{SPM-Tracker}}  & \tabincell{c}{SiamRPN++ \\ \cite{SiamRPN++}}  & \tabincell{c}{SiamMask \\ \cite{SiamMask}}  & \tabincell{c}{ARTCS \\ \cite{VOT2019}}  & \tabincell{c}{SiamDW\_ST \\ \cite{SiamDW}}  & \tabincell{c}{DCFST \\ \cite{VOT2019}}  & \tabincell{c}{DiMP \\ \cite{DiMP}}  & \textbf{Ours}  \\
\midrule
EAO$\uparrow$          & 0.252       & 0.262       & 0.275 & 0.285     & 0.287    & 0.287 & 0.299      & 0.317 & {\color{blue} 0.321} & {\color{red} 0.327} \\
Accuracy$\uparrow$     & 0.563       & 0.549       & 0.577 & 0.599     & 0.594    & {\color{red} 0.602} & 0.600      & 0.585 & 0.582 & {\color{red} 0.602} \\
Robustness$\downarrow$ & 0.507       & {\color{red} 0.346}       & 0.507 & 0.482     & 0.461    & 0.482 & 0.467      & 0.376 & {\color{blue} 0.371} & 0.396 \\
\bottomrule
\end{tabular}}
\end{center}
\caption{Detailed comparisons on VOT2019 real-time experiments.}
\label{tab:eao_rank_vot2019}
\end{table*}

\begin{figure}[t]
\begin{center}
  \includegraphics[width=1\linewidth]{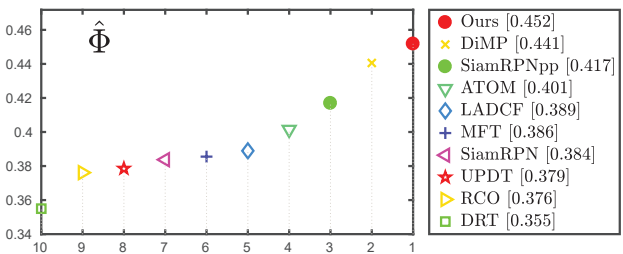}
\end{center}
\caption{Expected averaged overlap performance on VOT2018. SiamRPNpp is SiamRPN++, the same below.}
\label{fig:eao_rank_vot2018}
\end{figure}

\noindent
\textbf{VOT2018 \cite{VOT2018}.} \ We evaluate our tracker on the Visual Object Tracking challenge 2018 (VOT2018) consisting of 60 sequences. The overall performance of the tracker is evaluated using the EAO (Expected Average Overlap), which combines accuracy (average overlap during successful tracking periods) and robustness (failure rate). Table \ref{tab:eao_rank_vot2018} shows the comparison with almost all the top-performing trackers in the VOT2018. Among previous approaches, DiMP \cite{DiMP} achieves the best EAO and SiamRPN++ \cite{SiamRPN++} achieves the best accuracy, they all use ResNet-50 to extract feature. DiMP has the same accuracy as our tracker, and although its failure rate is slightly lower than ours, our EAO is slightly better, without any online update. Compared with SiamRPN++, our tracker achieves similar accuracy, but the failure rate decreases by 23.9\% and EAO increases by 8.4\%. Among these trackers, our tracker has the highest EAO and ranks second in terms of accuracy. This shows that our tracker not only accurately estimates the target's location but also maintain good robustness.

\begin{figure}[t]
\begin{center}
  \includegraphics[width=1\linewidth]{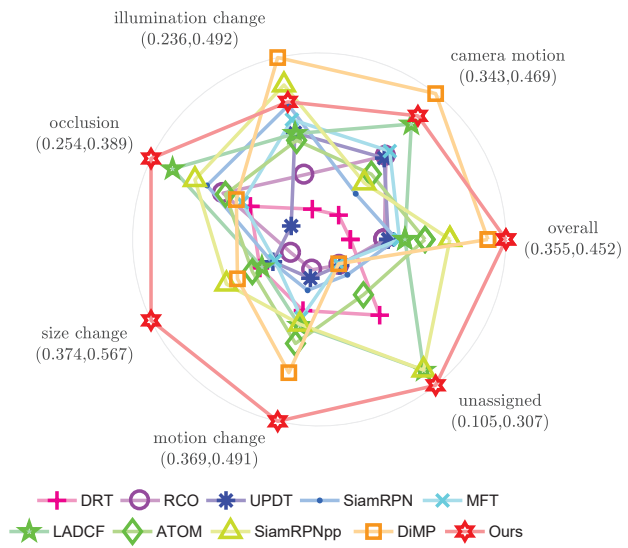}
\end{center}
\caption{Comparison of EAO on VOT2018 for the following visual attributes: camera motion, illumination change, occlusion, size change and motion change. Frames that do not correspond to any of the five attributes are marked as unassigned. The values in parentheses indicate the EAO range of each attribute and overall of the trackers.}
\label{fig:attr_eao_vot2018}
\end{figure}

\noindent
\textbf{Comparison of attributes on VOT2018.} \ All sequences of VOT2018 are per-frame annotated by the following visual attributes: camera motion, illumination change, occlusion, size change, and motion change. Frames that do not correspond to any of the five attributes are represented as unassigned. We compare the EAO of the visual attributes of the top-performing trackers. As shown in Figure \ref{fig:attr_eao_vot2018}, our tracker ranks first on attributes of occlusion, size change, and motion change, and ranks second and third on attributes of camera motion and illumination. This shows that our tracker is robust to occlusion, size changes, and motion changes in the target while having the ability to cope with camera motion and illumination changes.

\begin{figure}[t]
\begin{center}
  \includegraphics[width=1\linewidth]{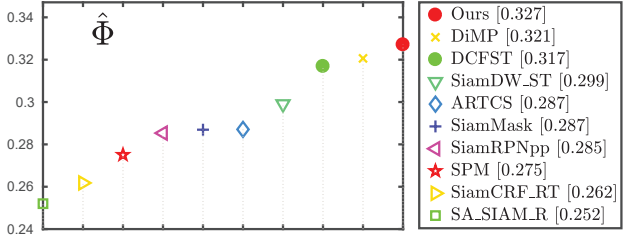}
\end{center}
\caption{Expected averaged overlap performance on VOT2019.}
\label{fig:eao_rank_vot2019}
\end{figure}

\noindent
\textbf{VOT2019 \cite{VOT2019}.} \ We evaluate our tracker on Visual Object Tracking challenge 2019 (VOT2019) real-time experiments. The VOT2019 sequences were replaced by 20\% compared to the VOT2018. Table \ref{tab:eao_rank_vot2019} shows the results presented in terms of EAO, robustness, and accuracy. SiamMargin \cite{VOT2019} achieves a lower failure rate through online updates, but our accuracy is higher than it. Although SiamRPN++ achieves similar accuracy to our tracker, our failure rate is 17.8\% lower than it and achieves 14.7\% relative gain in EAO. Among these trackers, our tracker has the highest accuracy and EAO. This shows that our method can accurately estimate the target bounding box.

\begin{figure}[t]
\begin{center}
\subfigure[Success Plot]{
\includegraphics[width=0.48\linewidth]{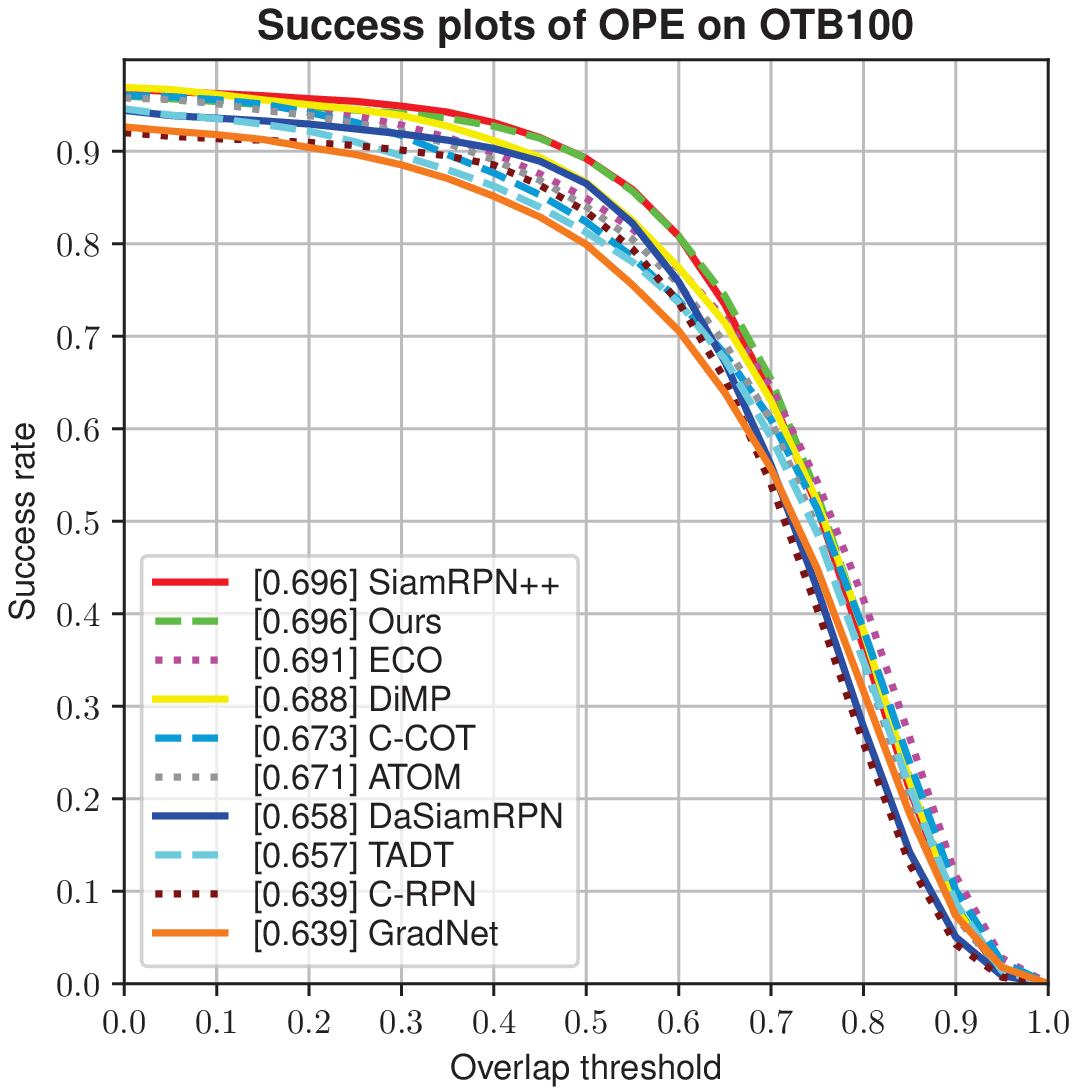}}
\subfigure[Precision Plot]{
\includegraphics[width=0.48\linewidth]{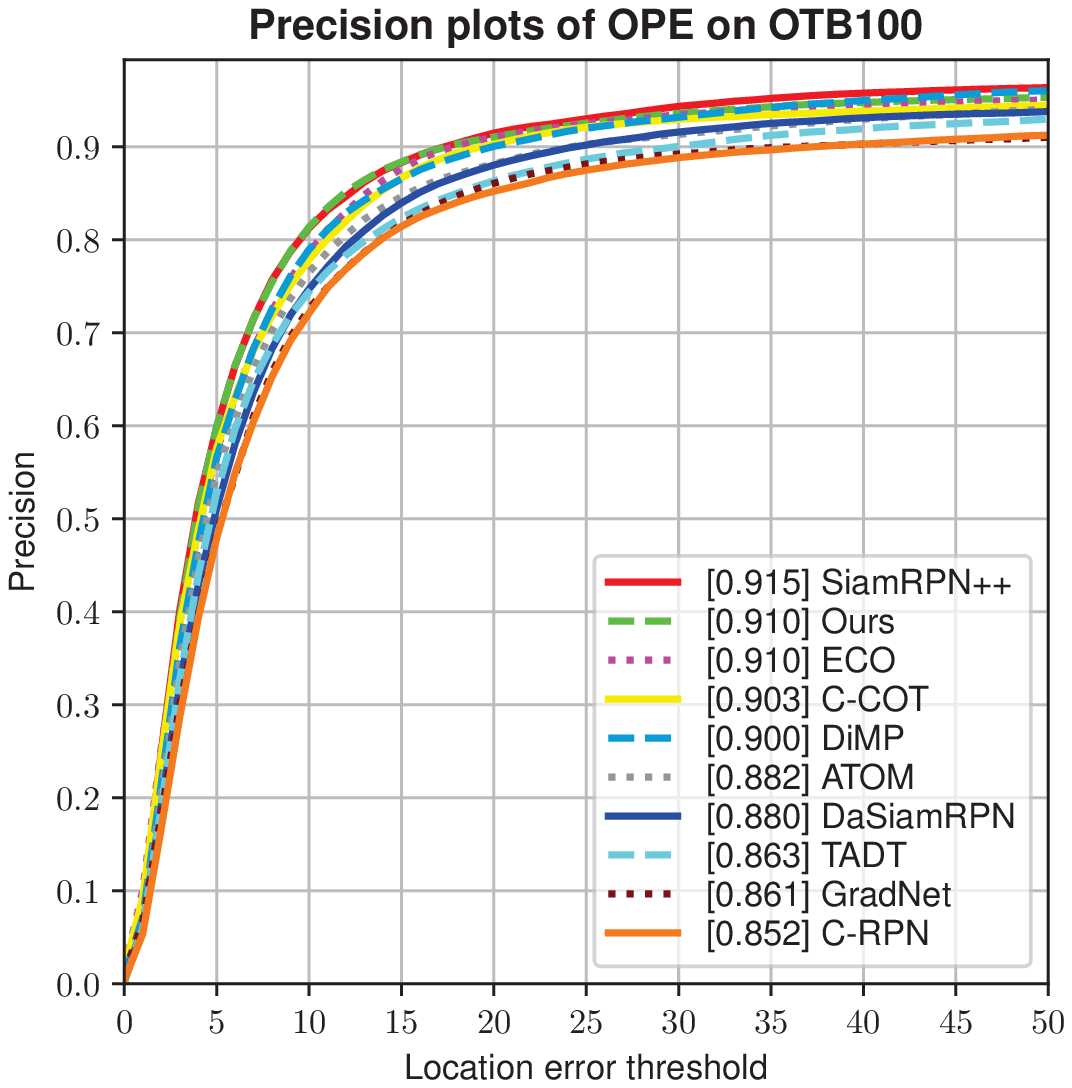}}
\end{center}
\caption{Success and precision plots on OTB100.}
\label{fig:otb100}
\end{figure}

\begin{figure}[t]
\begin{center}
\subfigure[Success Plot]{
\includegraphics[width=0.48\linewidth]{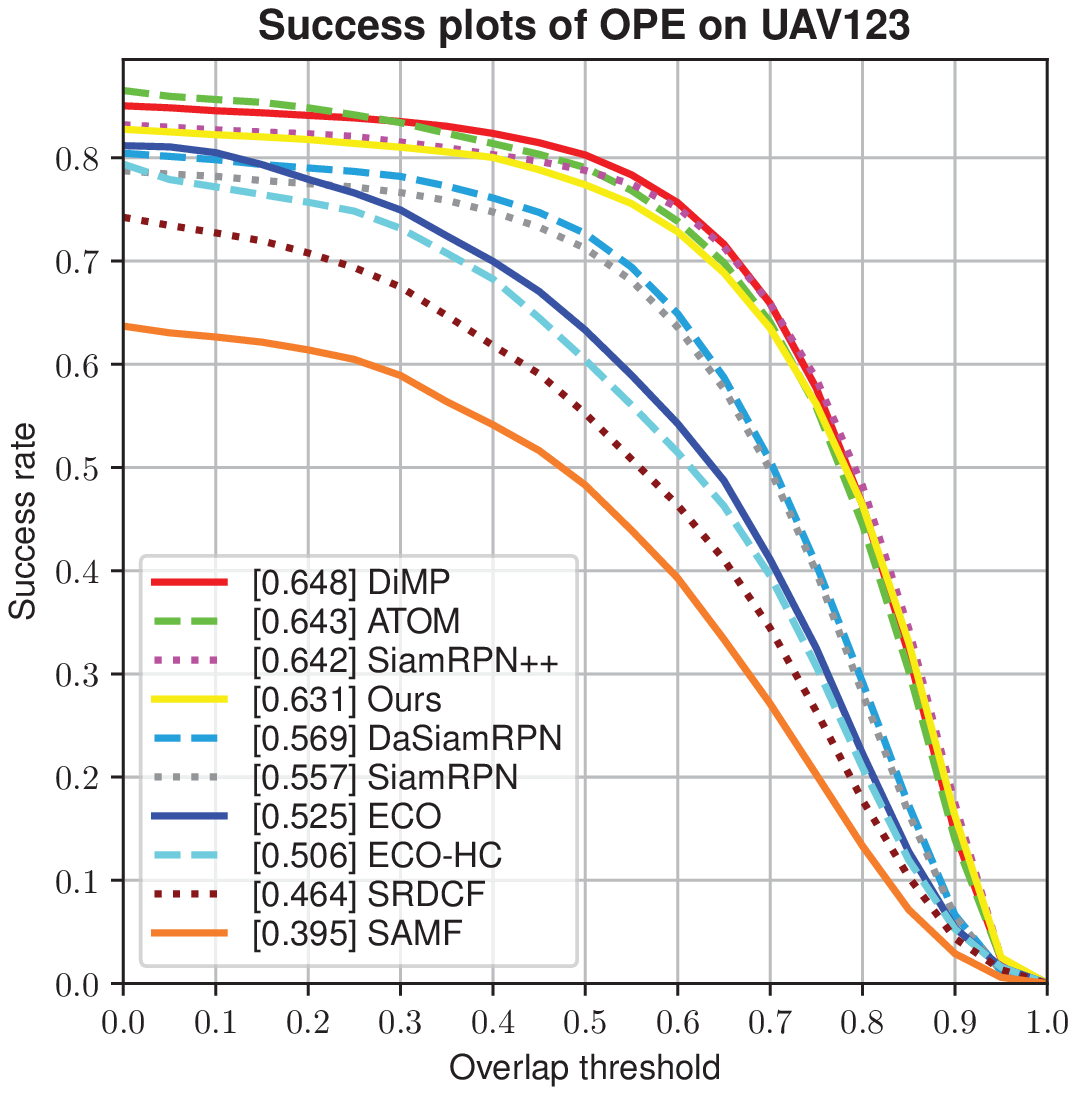}}
\subfigure[Precision Plot]{
\includegraphics[width=0.48\linewidth]{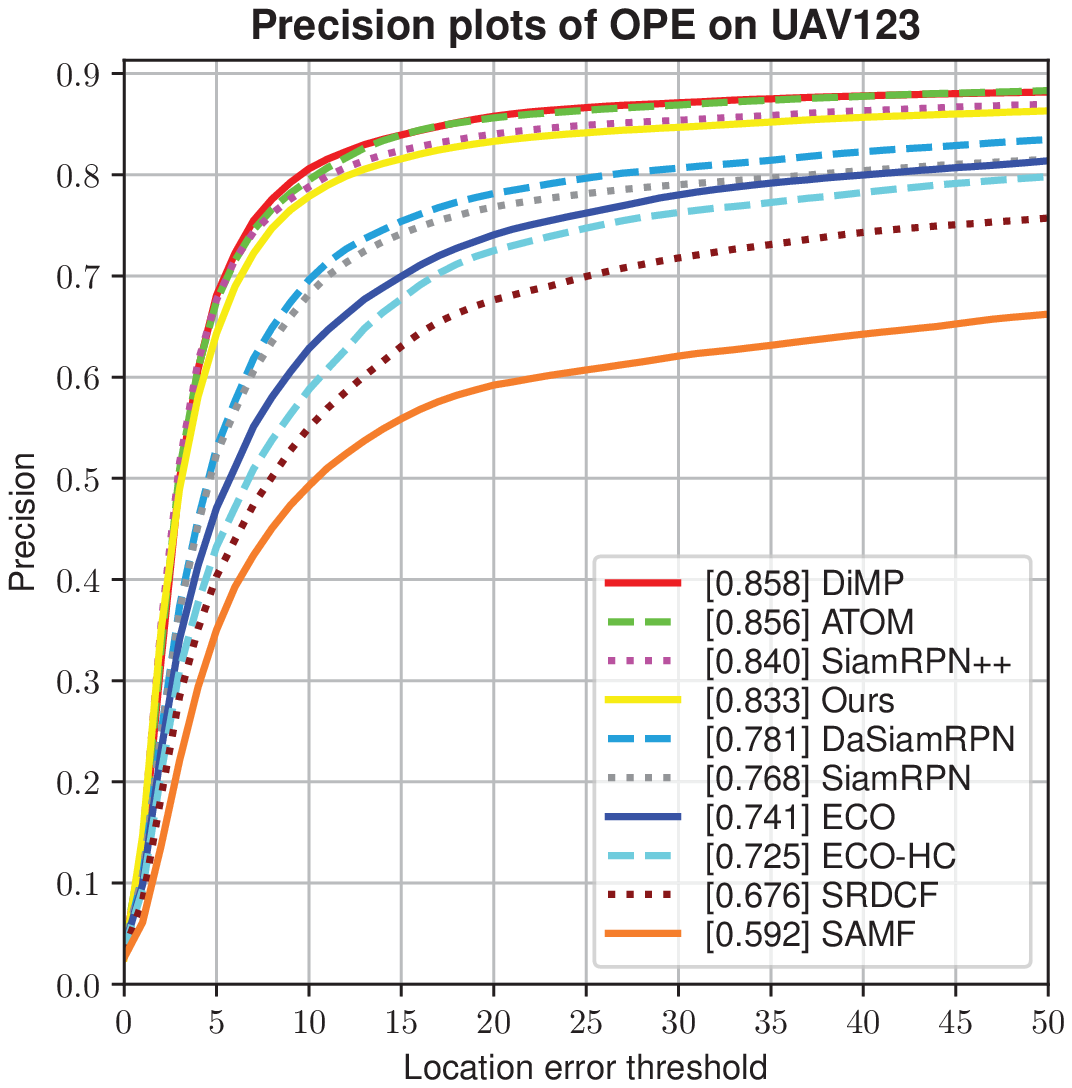}}
\end{center}
\caption{Success and precision plots on UAV123.}
\label{fig:uav123}
\end{figure}

\begin{figure}[t]
\begin{center}
\subfigure[Success Plot]{
\includegraphics[width=0.48\linewidth]{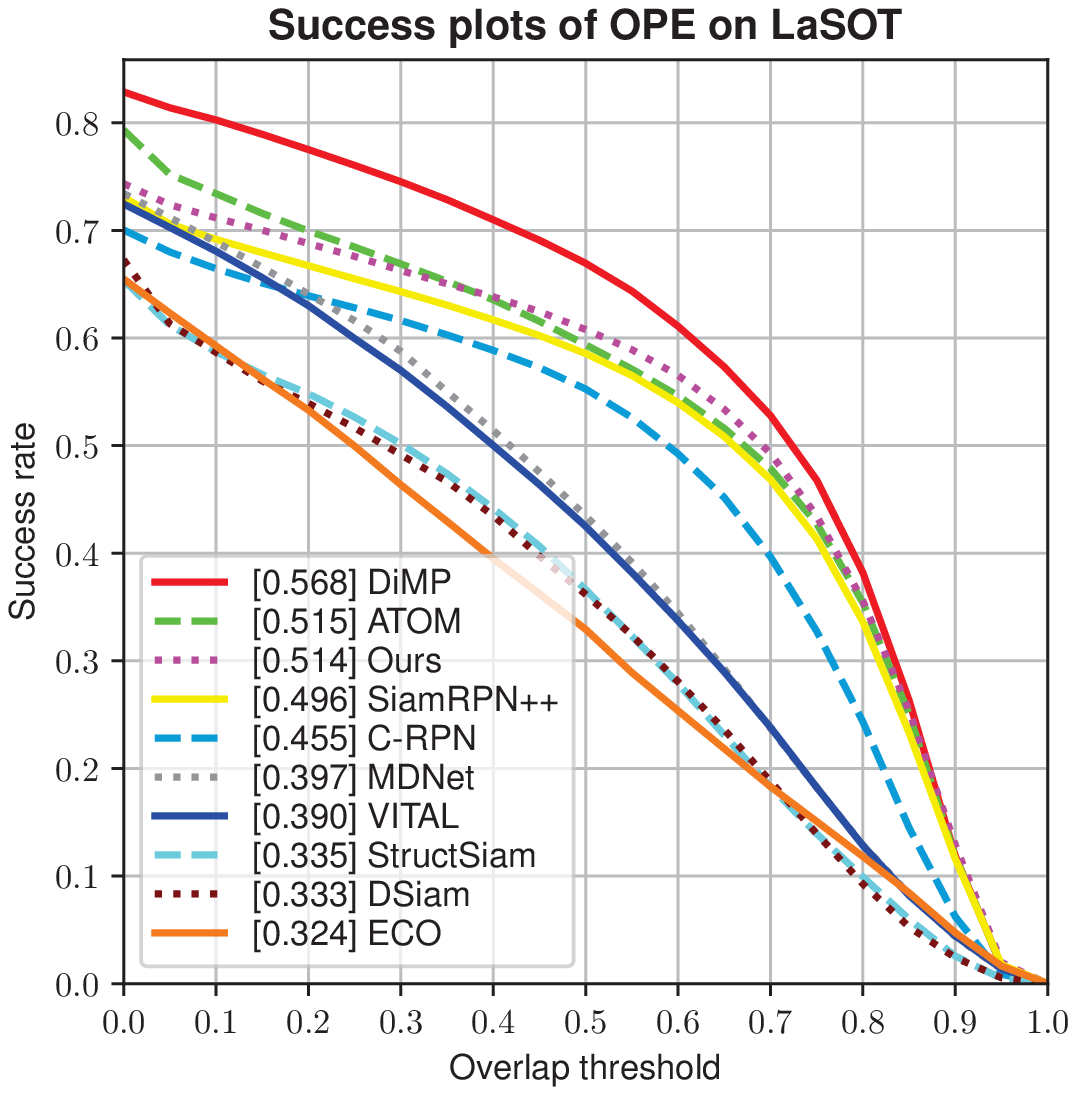}}
\subfigure[Normalized Precision Plot]{
\includegraphics[width=0.48\linewidth]{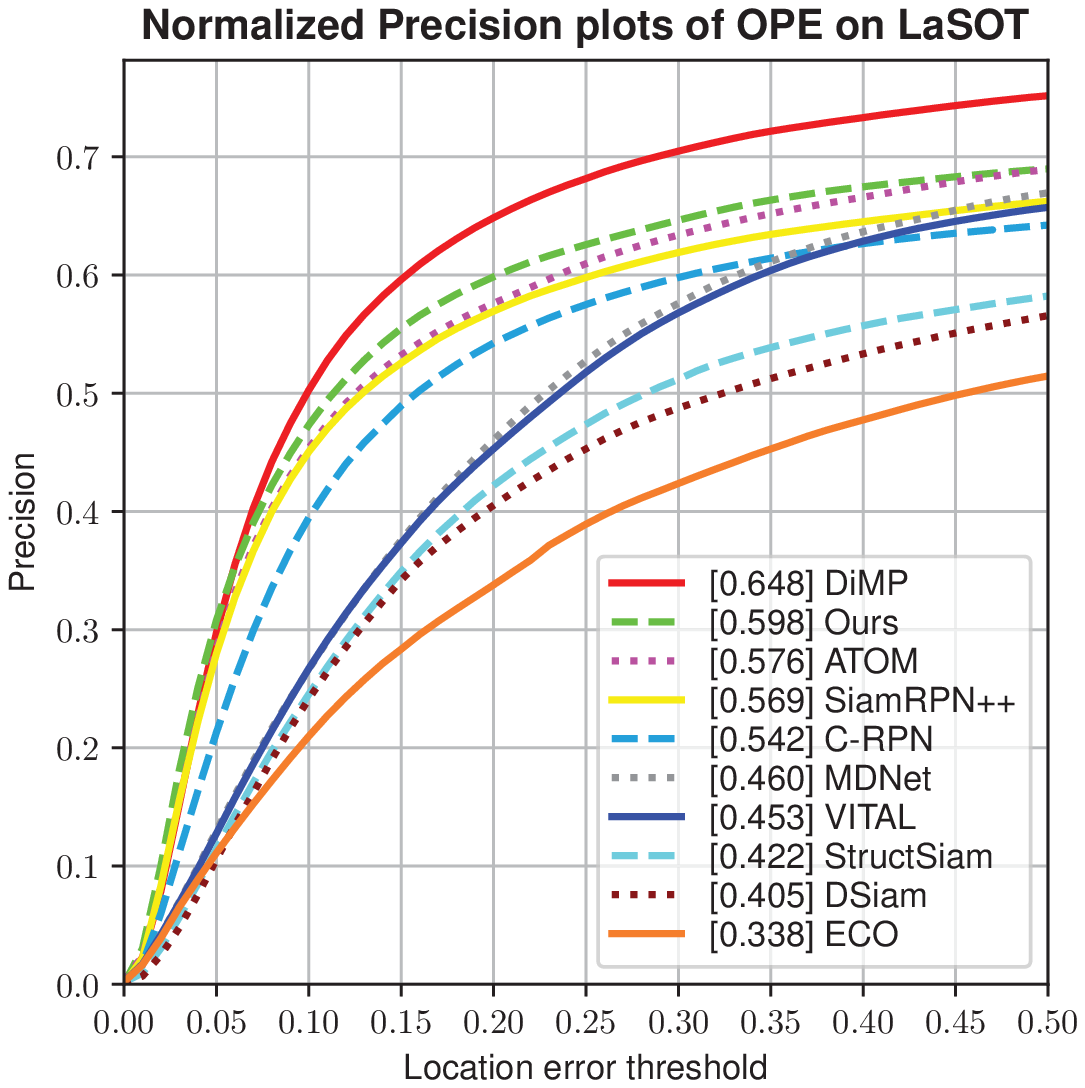}}
\end{center}
\caption{Success and normalized precision plots on LaSOT.}
\label{fig:lasot}
\end{figure}

\noindent
\textbf{OTB100 \cite{OTB100}.} \ OTB100 is a widely used public tracking benchmark consisting of 100 sequences. Our SiamBAN tracker is compared with numerous state-of-the-art trackers including SiamRPN++ \cite{SiamRPN++}, ECO \cite{ECO}, DiMP \cite{DiMP}, C-COT \cite{C-COT}, ATOM \cite{ATOM}, DaSiamRPN \cite{DaSiamRPN}, TADT \cite{TADT}, C-RPN \cite{C-RPN}, GradNet \cite{GradNet}. Figure \ref{fig:otb100} illustrates the success and precision plots of the compared trackers. Prior to SiamRPN++, due to the limited representation capabilities of shallow networks, the Siamese network based \cite{DaSiamRPN} trackers achieves sub-optimal performance on the OTB100. After using ResNet-50 as the feature extraction network, SiamRPN++ achieves leading results. Compared to SiamRPN++, achieves similar results with a simpler design.

\begin{table}[t]
\begin{center}
\setlength{\tabcolsep}{0.5mm}{
\begin{tabular}{lccccccc}
\toprule
& \tabincell{c}{MDNet \\ \cite{MDNet}}  & \tabincell{c}{ECO \\ \cite{ECO}}  & \tabincell{c}{C-COT \\ \cite{C-COT}} & \tabincell{c}{UPDT \\ \cite{UPDT}}  & \tabincell{c}{ATOM \\ \cite{ATOM}}  & \tabincell{c}{DiMP \\ \cite{DiMP}}  & \textbf{Ours}  \\
\midrule
AUC$\uparrow$ & 0.422 & 0.466 & 0.488 & 0.537 & 0.584 & {\color{red} 0.620 } & {\color{blue} 0.594 } \\
\bottomrule
\end{tabular}}
\end{center}
\caption{Comparison with State-of-the-art trackers on the NFS datase in terms of AUC.}
\label{tab:nfs}
\end{table}

\noindent
\textbf{NFS \cite{NFS}.} \ The NFS dataset consists of 100 videos (380K frames) captured from real-world scenes with higher frame rate cameras. We evaluate our tracker in the 30FPS version of the dataset, AUC are shown in Table \ref{tab:nfs}. It can be seen that our tracker ranks second and improved by 40.8\% compared to the best tracker in the NFS paper.

\noindent
\textbf{UAV123 \cite{UAV123}.} \ The UAV123 is a new aerial video benchmark and dataset, which contains 123 sequences captured from a low-altitude aerial perspective. The benchmarks can be used to assess whether the tracker is suitable for deployment to a UAV in real-time scenarios. We compare our tracker with other 9 state-of-art real-time trackers, including DiMP \cite{DiMP}, ATOM \cite{ATOM}, SiamRPN++ \cite{SiamRPN++}, DaSiamRPN \cite{DaSiamRPN}, SiamRPN \cite{SiamRPN}, ECO \cite{ECO}, ECO-HC \cite{ECO}, SRDCF \cite{SRDCF}, SAMF \cite{SAMF}. Figure \ref{fig:uav123} shows the success and precision plots. Our tracker achieves state-of-the-art score.

\noindent
\textbf{LaSOT \cite{LaSOT}.} \ LaSOT is a high-quality, large-scale dataset with a total of 1,400 sequences. Compared to the previous dataset, LaSOT has longer sequences with an average sequence length of more than 2,500 frames. Each sequence has various challenges from the wild where the target may disappear and reappear in the view, which tests the ability of the tracker to re-track the target. We evaluate our tracker on the test set consisting of 280 videos with trackers including DiMP \cite{DiMP}, ATOM \cite{ATOM}, SiamRPN++ \cite{SiamRPN++}, C-RPN \cite{C-RPN}, MDNet \cite{MDNet}, VITAL \cite{VITAL}, StructSiam \cite{StructSiam}, DSiam\cite{DSiam}, ECO \cite{ECO}. The results including success plots and normalized precision plots are illustrated in Figure \ref{fig:lasot}. Our tracker ranks third in terms of AUC, second in terms of normalized precision and 5.1\% higher than SiamRPN++. 

\subsection{Ablation Study}
\noindent
\textbf{Discussion on Multi-level Prediction.} \ To explore the role of different level features and the effect of aggregation of multi-level features, we have performed an ablation study on multi-layer prediction. It can be found from Table \ref{tab:ablation_study} that when only single-layer feature are used, $conv4$ performs best. Compared with the single-layer features, when using the aggregation of the two-layer features, the performance has been improved, and the performance of $conv4$ and $conv5$ aggregation is the best. After aggregating three layers of features, our tracker achieves the best results.

\noindent
\textbf{Discussion on Sample Label Assignment.} \ The sample label assignment plays a key role in the performance of a tracker. However, many Siamese network based trackers \cite{SiamFC, CFNet, DSiam} do not pay enough attention to it. For example, SiamFC considers the elements of the score map within the radius R of the center to be positive samples. The label assignment method only considers the center position of the target, ignoring the size of the target. Intuitively, the sample label assignment should be different for targets of different sizes and shapes. Therefore, our label assignment also takes into account the target scale and aspect ratio. It is worth noting that we also set the buffer to ignore the ambiguous samples. The specific is in Section \ref{sec:ground-truth and loss}. 

\begin{figure}[t]
\begin{center}
  \includegraphics[width=1\linewidth]{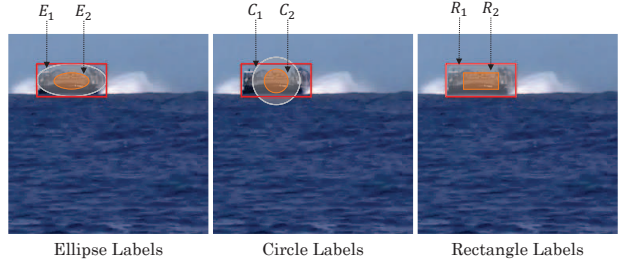}
\end{center}
\caption{Three sample label assignment methods: ellipse labels, circle labels, rectangular labels. $E_1$, $E_2$, $C_1$, $C_2$, $R_1$, $R_2$ represent ellipse $E_1$, ellipse $E_2$, circle $C_1$, circle $C_2$, rectangle $R_1$, rectangle $R_2$, respectively.}
\label{fig:label_assignment}
\end{figure}

To illustrate the advantages of our label assignment method, we conduct comparative experiments with the other two label assignments. As shown in Figure \ref{fig:label_assignment}, for convenience, we refer to these three types of labels as ellipse labels, circle labels and rectangle labels. For fair comparison, we define circles $C_1$, $C_2$ and rectangles $R_1$, $R_2$ in a similar way to define ellipses $E_2$, $E_2$. Specifically, with $(g_{x_c},\ g_{y_c})$ as the center and $\frac{\sqrt{g_{w} \times g_{h}}}{2}$, $\frac{\sqrt{g_{w} \times g_{h}}}{4}$ as the radius, we can get the circles $C_1$ and $C_2$. The rectangle $R_1$ is the same position and size as the ground-truth bounding box. The center of the rectangle $R_2$ is $(g_{x_c},\ g_{y_c})$, and the sides length is $\frac{g_w}{2}$, $\frac{g_h}{2}$.

As shown in Table \ref{tab:ablation_study}, under the same number of iterations and training dataset, SiamBAN performs better than SiamBAN with circle labels and SiamBAN with rectangle labels. We believe that the reason is that ellipse labels can more accurately label positive and negative samples than circular labels and rectangular labels so that the trained tracker can more accurately distinguish the foreground-background and is more robust.


\begin{table}[t]
\begin{center}
\begin{tabular}{ccccccc}
\toprule
L3           & L4           & L5           & Circle       & Rectangle    & Ellipse      & AUC   \\
\midrule
$\checkmark$ &              &              &              &              & $\checkmark$ & 0.675 \\
             & $\checkmark$ &              &              &              & $\checkmark$ & 0.683 \\
             &              & $\checkmark$ &              &              & $\checkmark$ & 0.662 \\
$\checkmark$ & $\checkmark$ &              &              &              & $\checkmark$ & 0.687 \\
$\checkmark$ &              & $\checkmark$ &              &              & $\checkmark$ & 0.681 \\
             & $\checkmark$ & $\checkmark$ &              &              & $\checkmark$ & 0.689 \\
\midrule
$\checkmark$ & $\checkmark$ & $\checkmark$ & $\checkmark$ &              &              & 0.686 \\
$\checkmark$ & $\checkmark$ & $\checkmark$ &              & $\checkmark$ &              & 0.688 \\
$\checkmark$ & $\checkmark$ & $\checkmark$ &              &              & $\checkmark$ & \color{red} 0.696 \\
\bottomrule
\end{tabular}
\end{center}
\caption{Quantitative comparison results of our tracker and its variants with different detection heads and different label assignment methods on OTB100. L3, L4, L5 represent $conv3$, $conv4$, $conv5$, respectively. Circle, Rectangle, Ellipse represent circle labels, rectangle labels, ellipse labels, respectively.}
\label{tab:ablation_study}
\end{table}

\section{Conclusions}

In this paper, we exploit the expressive power of the fully convolutional network and propose a simple yet effective visual tracking framework named SiamBAN, which does not require a multi-scale searching schema and pre-defined candidate boxes. SiamBAN directly classifies objects and regresses bounding boxes in a unified network. Therefore, the visual tracking problem becomes a classification-regression problem. Extensive experiments on six visual tracking benchmarks demonstrate that SiamBAN achieves state-of-the-art performance and runs at 40 FPS, confirming its effectiveness and efficiency.

\noindent
\textbf{Acknowledgments.}
This work was supported by the National Natural Science Foundation of China (No. 61972167, 61772494, 61872112), the Fundamental Research Funds for the Central Universities (No. 30918014108), and the Open Project Program of the National Laboratory of Pattern Recognition (NLPR) (No. 202000012).

{\small
\bibliographystyle{ieee_fullname}
\bibliography{egbib}
}

\end{document}